\documentclass[runningheads,a4paper]{llncs}

\usepackage[misc]{ifsym}
\usepackage{amssymb}
\usepackage{graphicx}
\usepackage{color}
\usepackage{url}
\usepackage[table]{xcolor}
\usepackage{multirow}
\usepackage{hyperref}
\usepackage{subfig}
\usepackage{todonotes}
\usepackage{amssymb}

\urldef{\mail}\path|j.m.wolterink@umcutrecht.nl|

\newcommand{\keywords}[1]{\par\addvspace\baselineskip
\noindent\keywordname\enspace\ignorespaces#1}

\begin{document}

\mainmatter  

\title{Deep MR to CT Synthesis using Unpaired Data}
\titlerunning{MR to CT Synthesis using Unpaired Training Data}

\author{Jelmer M. Wolterink\inst{1}\Letter, Anna M. Dinkla\inst{2}, Mark H.F. Savenije\inst{2}, \\ Peter R. Seevinck\inst{1}, Cornelis A.T. van den Berg\inst{2}, Ivana I\v{s}gum\inst{1}}


\authorrunning{J.M. Wolterink et al.}

\institute{Image Sciences Institute, University Medical Center Utrecht, The Netherlands \\ \mail \and Department of Radiotherapy, University Medical Center Utrecht, The Netherlands}

\maketitle

\begin{abstract}
MR-only radiotherapy treatment planning requires accurate MR-to-CT synthesis. Current deep learning methods for MR-to-CT synthesis depend on pairwise aligned MR and CT training images of the same patient. However, misalignment between paired images could lead to errors in synthesized CT images. To overcome this, we propose to train a generative adversarial network (GAN) with unpaired MR and CT images. A GAN consisting of two synthesis convolutional neural networks (CNNs) and two discriminator CNNs was trained with cycle consistency to transform 2D brain MR image slices into 2D brain CT image slices and vice versa. Brain MR and CT images of 24 patients were analyzed.  A quantitative evaluation showed that the model was able to synthesize CT images that closely approximate reference CT images, and was able to outperform a GAN model trained with paired MR and CT images. 
\end{abstract}

\section{Introduction}
Radiotherapy treatment planning requires a magnetic resonance (MR) volume for segmentation of tumor volume and organs at risk, as well as a spatially corresponding computed tomography (CT) volume for dose planning. Separate acquisition of these volumes is time-consuming, costly and a burden to the patient. Furthermore, voxel-wise spatial alignment between MR and CT images may be compromised, requiring accurate registration of MR and CT volumes. 
Hence, to circumvent separate CT acquisition, a range of methods have been proposed for MR-only radiotherapy treatment planning in which a substitute or synthetic CT image is derived from the available MR image \cite{Edmu17}. 

Previously proposed methods have used convolutional neural networks (CNNs) for CT synthesis in the brain \cite{Han17} and pelvic area \cite{Nie16}. These CNNs are trained by minimization of voxel-wise differences with respect to reference CT volumes that are rigidly aligned with the input MR images. 
However, slight voxel-wise misalignment of MR and CT images may lead to synthesis of blurred images. 
To address this, Nie et al. \cite{Nie16b} proposed to combine the voxel-wise loss with an image-wise adversarial loss in a generative adversarial network (GAN) \cite{Good14}. In this GAN, the synthesis CNN competes with a discriminator CNN that aims to distinguish synthetic images from real CT images. The discriminator CNN provides feedback to the synthesis CNN based on the overall quality of the synthesized CT images.

\begin{figure}[t!]
\centering
\includegraphics[width=0.9\textwidth]{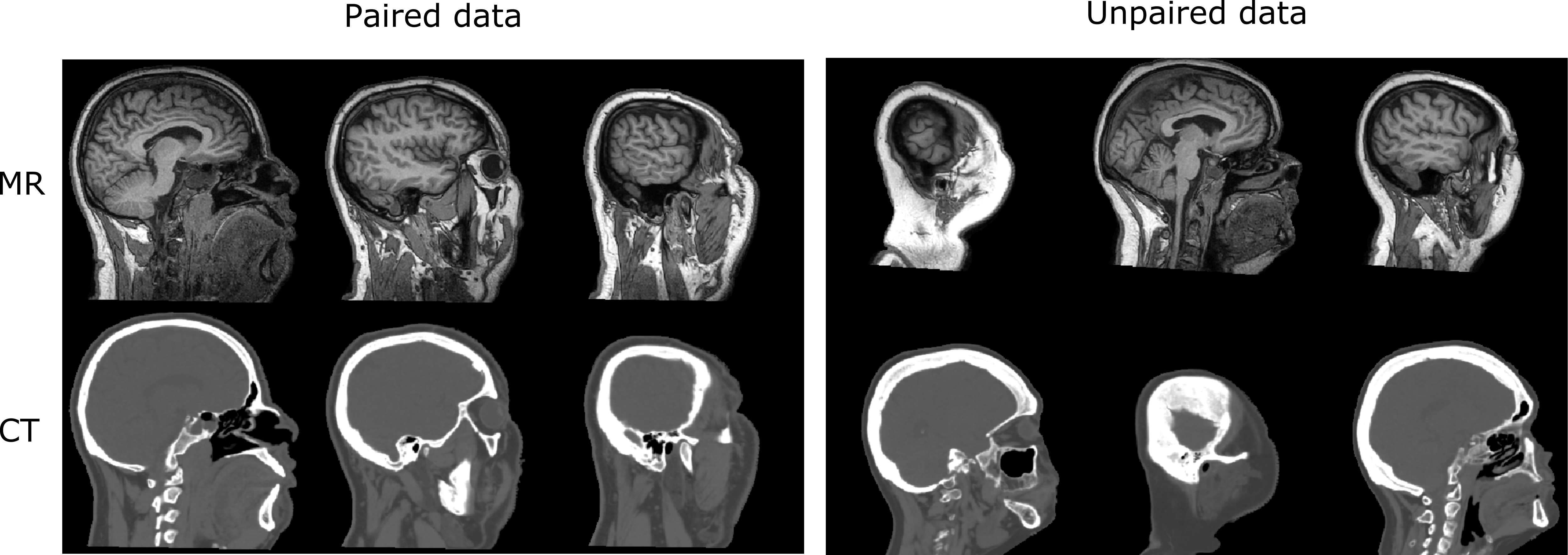}
\caption{\textit{Left} When training with paired data, MR and CT slices that are simultaneously provided to the network correspond to the same patient at the same anatomical location. \textit{Right} When training with unpaired data, MR and CT slices that are simultaneously provided to the network belong to different patients at different locations in the brain.}
\label{fig:examples}
\end{figure}

Although the GAN method by Nie et al. \cite{Nie16b} addresses the issue of image misalignment by incorporating an image-wise loss, it still contains a voxel-wise loss component requiring a training set of paired MR and CT volumes. In practice, such a training set may be hard to obtain. 
Furthermore, given the scarcity of training data, it may be beneficial to utilize additional MR or CT training volumes from patients who were scanned for different purposes and who have not necessarily been imaged using both modalities. The use of unpaired MR and CT training data would relax many of the requirements of current deep learning-based CT synthesis systems (Fig. \ref{fig:examples}).

Recently, methods have been proposed to train image-to-image translation CNNs with unpaired natural images, namely DualGAN \cite{Yi17} and CycleGAN \cite{Zhu17}. Like the methods proposed in \cite{Han17,Nie16,Nie16b}, these CNNs translate an image from one domain to another domain. Unlike these methods, the loss function during training depends solely on the overall quality of the synthesized image as determined by an adversarial discriminator network. 
To prevent the synthesis CNN from generating images that look real but bear little similarity to the input image, cycle consistency is enforced. That is, an additional CNN is trained to translate the synthesized image back to the original domain and the difference between this reconstructed image and the original image is added as a regularization term during training.

Here, we use a CycleGAN model to synthesize brain CT images from brain MR images. We show that training with pairs of spatially aligned MR and CT images of the same patients is not necessary for deep learning-based CT synthesis.

\section{Data}
\begin{figure}[t!]
\centering
\includegraphics[width=0.9\textwidth]{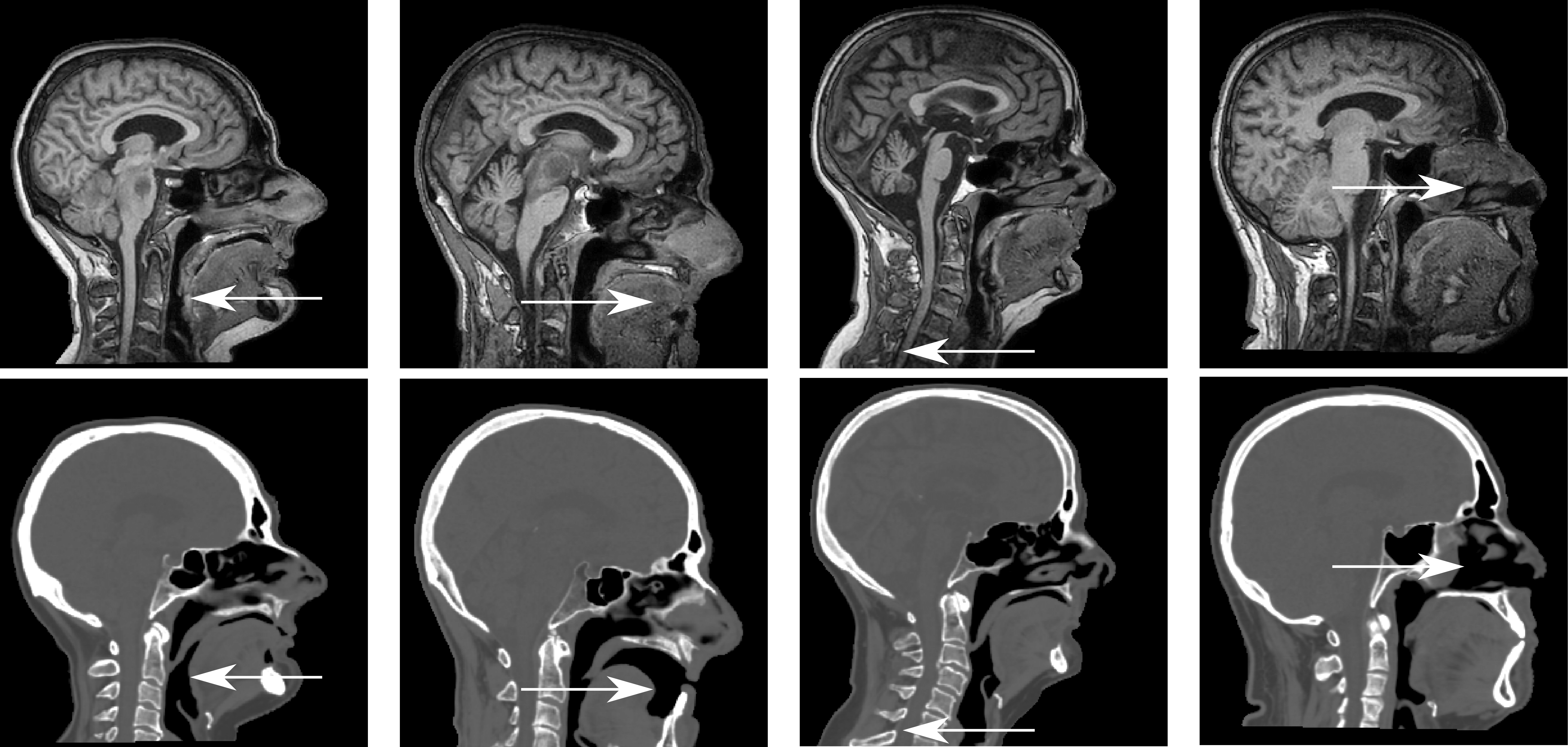}
\caption{Examples showing local misalignment between MR and CT images after rigid registration using mutual information. Although the skull is generally well-aligned, misalignments may occur in the throat, mouth, vertebrae, and nasal cavities.} 
\label{fig:misalignment}
\end{figure}

This study included brain MR and CT images of 24 patients that were scanned for radiotherapy treatment planning of brain tumors. MR and CT images were acquired on the same day in radiation treatment position using a thermoplastic mask for immobilization. Patients with heavy dental artefacts on CT and/or MR were excluded.
T1 3D MR (repetition time 6.98 ms, echo time 3.14 ms, flip angle 8$^{\circ}$) images were obtained with dual flex coils on a Philips Ingenia 1.5T MR scanner (Philips Healthcare, Best, The Netherlands). CT images were acquired helically on a Philips Brilliance Big Bore CT scanner (Philips Healthcare, Best, The Netherlands) with 120 kVp and 450 mAs. 
To allow voxel-wise comparison of synthetic and reference CT images, MR and CT images of the same patient were aligned using rigid registration based on mutual information following a clinical procedure. This registration did not correct for local misalignment (Fig. \ref{fig:misalignment}). CT images had a resolution of $1.00\times 0.90\times 0.90$ mm$^3$ and were resampled to the same voxel size as the MR, namely $1.00\times 0.87\times 0.87$ mm$^3$. Each volume had $183\times 288\times 288$ voxels. A head region mask excluding surrounding air was obtained in the CT image and propagated to the MR image.

\section{Methods}
\begin{figure}[tp!]
\centering
\includegraphics[width=1.0\textwidth]{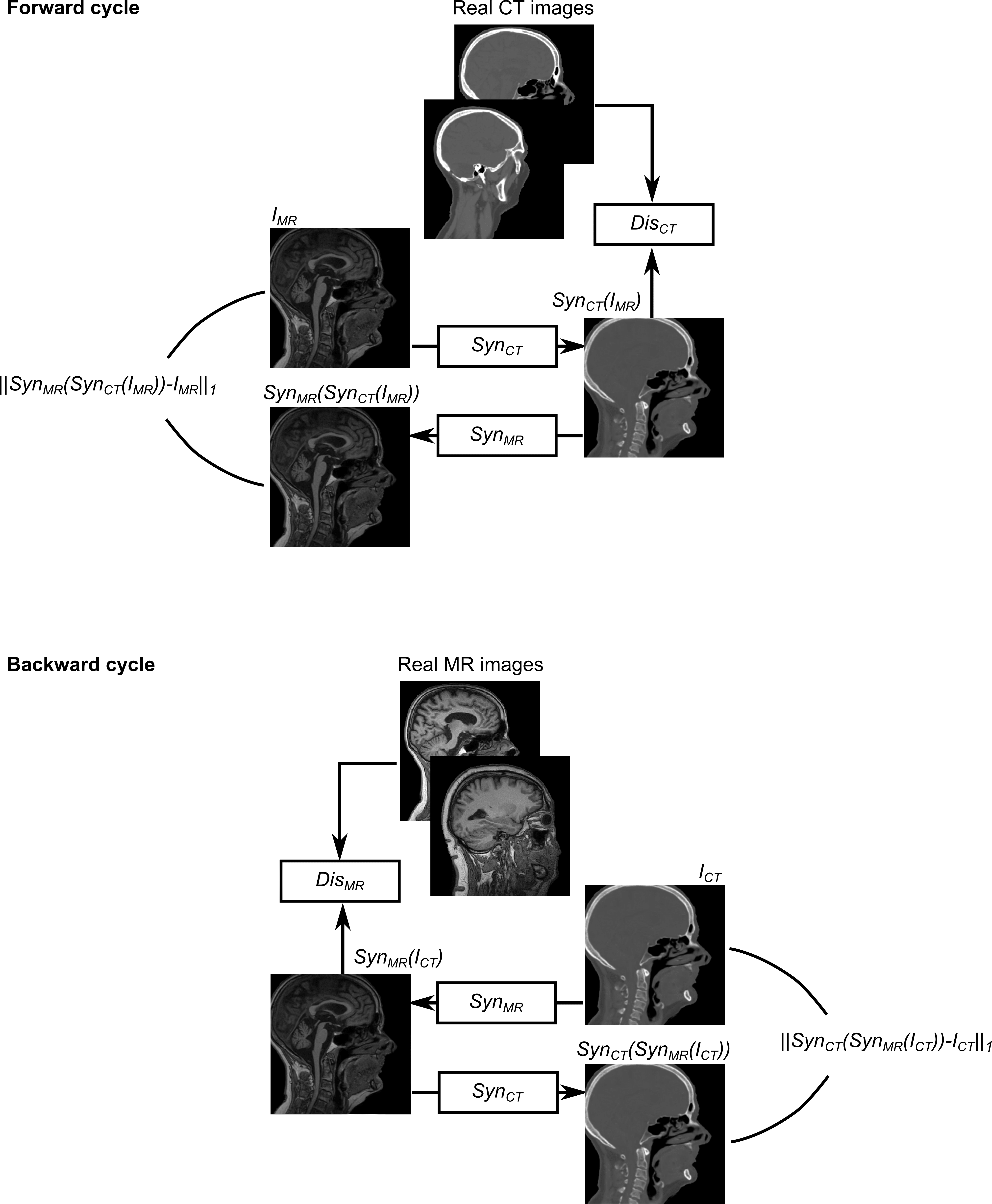}
\caption{The CycleGAN model consists of a forward cycle and a backward cycle. In the forward cycle, a synthesis network $Syn_{CT}$ is trained to translate an input MR image $I_{MR}$ into a CT image, network $Syn_{MR}$ is trained to translate the resulting CT image back into an MR image that approximates the original MR image, and $Dis_{CT}$ discriminates between real and synthesized CT images. In the backward cycle, $Syn_{MR}$ synthesizes MR images from input CT images, $Syn_{CT}$ reconstructs the input CT image from the synthesized image, and $Dis_{MR}$ discriminates between real and synthesized MR images.} 
\label{fig:cycleconsistency}
\end{figure}

The CycleGAN model proposed by Zhu et al. and used in this work contains a forward and a backward cycle (Fig. \ref{fig:cycleconsistency}) \cite{Zhu17}. 

The forward cycle consists of three separate CNNs. First, network $Syn_{CT}$ is trained to translate an input MR image $I_{MR}$ into a CT image. Second, network $Syn_{MR}$ is trained to translate a synthesized CT image $Syn_{CT}(I_{MR})$ back into an MR image. Third, network $Dis_{CT}$ is trained to discriminate between synthesized $Syn_{CT}(I_{MR})$ and real CT images $I_{CT}$. Each of these three neural networks has a different goal. While $Dis_{CT}$ aims to distinguish synthesized CT images from real CT images, network $Syn_{CT}$ tries to prevent this by synthesizing images that cannot be distinguished from real CT images. These images should be translated back to the MR domain by network $Syn_{MR}$ so that the original image is reconstructed from $Syn_{CT}(I_{MR})$ as accurately as possible.

To improve training stability, the backward cycle is also trained, translating CT images into MR images and back into CT images. For synthesis, this model uses the same CNNs $Syn_{CT}$ and $Syn_{MR}$. In addition, it contains a discriminator network $Dis_{MR}$ that aims to distinguish synthesized MR images from real MR images. 

The adversarial goals of the synthesis and discriminator networks are reflected in their loss functions. The discriminator $Dis_{CT}$ aims to predict the label $1$ for real CT images and the label $0$ for synthesized CT images. Hence, the discriminator $Dis_{CT}$ tries to minimize

\begin{equation}
\mathcal{L}_{CT} = (1-Dis_{CT}(I_{CT}))^2 + Dis_{CT}(Syn_{CT}(I_{MR}))^2
\end{equation}

for MR images $I_{MR}$ and CT images $I_{CT}$. At the same time, synthesis network $Syn_{CT}$ tries to maximize this loss by synthesizing images that cannot be distinguished from real CT images. 

Similarly, the discriminator $Dis_{MR}$ aims to predict the label $1$ for real MR images and the label $0$ for synthesized MR images. Hence, the loss function for MR synthesis that $Dis_{MR}$ aims to minimize and $Syn_{MR}$ aims to maximize is defined as

\begin{equation}
\mathcal{L}_{MR} = (1-Dis_{MR}(I_{MR}))^2 + Dis_{MR}(Syn_{MR}(I_{CT}))^2
\end{equation}

To enforce bidirectional cycle consistency during training, additional loss terms are defined as the difference between original and reconstructed images, 

\begin{equation}
\mathcal{L}_{Cycle} = ||Syn_{MR}(Syn_{CT}(I_{MR}))-I_{MR}||_1 + ||Syn_{CT}(Syn_{MR}(I_{CT}))-I_{CT}||_1.
\end{equation}

During training, this term is weighted by a parameter $\lambda$ and added to the loss functions for $Syn_{CT}$ and $Syn_{MR}$.

\subsection{CNN Architectures}

The PyTorch implementation provided by the authors of \cite{Zhu17} was used in all experiments\footnote{\url{https://github.com/junyanz/pytorch-CycleGAN-and-pix2pix}}. 
This implementation performs voxel regression and image classification in 2D images. Here, experiments were performed using 2D sagittal image slices (Fig. \ref{fig:examples}). We provide a brief description of the synthesis and discriminator CNNs. Further implementation details are provided in \cite{Zhu17}. 

The network architectures of $Syn_{CT}$ and $Syn_{MR}$ are identical. They are 2D fully convolutional networks with two strided convolution layers, nine residual blocks and two fractionally strided convolution layers, based on the architecture proposed in \cite{John16} and used in \cite{Zhu17}. Hence, the CNN takes input images of size $256\times 256$ pixels and predicts output images of the same size.

Networks $Dis_{CT}$ and $Dis_{MR}$ also use the same architecture. This architecture does not provide one prediction for the full $256\times 256$ pixel image, but instead uses a fully convolutional architecture to classify overlapping $70\times 70$ image patches as real or fake \cite{Isol16}. This way, the CNN can better focus on high-frequency information that may distinguish real from synthesized images.

\subsection{Evaluation}
Real and synthesized CT images were compared using the mean absolute error

\begin{equation}
MAE = \frac{1}{N}\sum_{i=1}^N|I_{CT}(i)-Syn_{CT}(I_{MR}(i))|,
\end{equation}

where $i$ iterates over aligned voxels in the real and synthesized CT images. Note that this was based on the prior alignment of $I_{MR}$ and $I_{CT}$.
In addition, agreement was evaluated using the peak-signal-to-noise-ratio (PSNR) as proposed in \cite{Nie16,Nie16b} as

\begin{equation}
PSNR = 20\log_{10}\frac{4095}{MSE}, 
\end{equation}

where $MSE$ is the mean-squared error, i.e. $\frac{1}{N}\sum_{i=1}^N(I_{CT}(i)-Syn_{CT}(I_{MR}(i))^2$.
The MAE and PSNR were computed within a head region mask determined in both the CT and MR that excludes any surrounding air.

\section{Experiments and Results}
The 24 data sets were separated into a training set containing MR and CT volumes of 18 patients and a separate test set containing MR and corresponding reference CT volumes of 6 patients. 

Each MR or CT volume contained 183 sagittal 2D image slices. 
These were resampled to $256\times 256$ pixel images with 256 grayscale values uniformly distributed in $[-600, 1400]$ HU for CT and $[0, 3500]$ for MR. This put image values in the same range as in \cite{Zhu17}, so that the default value of $\lambda=10$ was used to weigh cycle consistency loss. To augment the number of training samples, each image was padded to $286\times 286$ pixels and sub-images of $256\times 256$ pixels were randomly cropped during training. 
The model was trained using Adam \cite{King15} for 100 epochs with a fixed learning rate of 0.0002, and 100 epochs in which the learning rate was linearly reduced to zero. Model training took 52 hours on a single NVIDIA Titan X GPU. MR to CT synthesis with a trained model took around 10 seconds.

\begin{figure}[tp!]
\includegraphics[width=\textwidth]{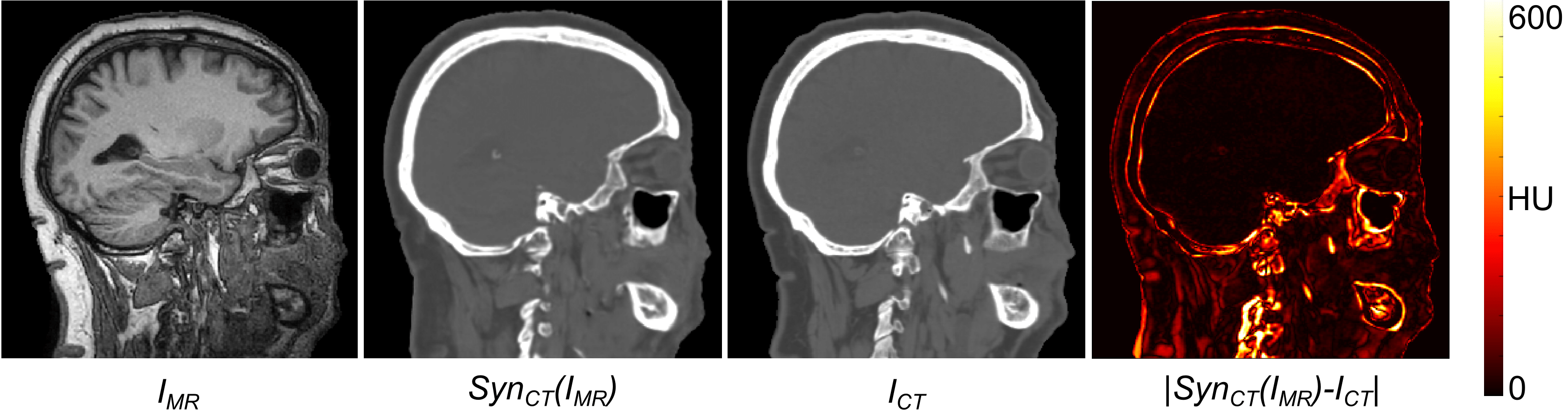}
\caption{\textit{From left to right} Input MR image, synthesized CT image, reference real CT image, and absolute error between real and synthesized CT image.}
\label{fig:resultmaps}
\end{figure}

Figure \ref{fig:resultmaps} shows an example MR input image, the synthesized CT image obtained by the model and the corresponding reference CT image. The model has learned to differentiate between different structures with similar intensity values in MR but not in CT, such as bone, ventricular fluid and air. The difference image shows the absolute error between the synthesized and real CT image. Differences are least pronounced in the soft brain tissue, and most in bone structures, such as the eye socket, the vertebrae and the jaw. This may be partly due to the reduced image quality in the neck area and misalignment between the MR image and the reference CT image. Table \ref{tab:maevalues} shows a quantitative comparison between real CT and synthesized CT images in the test set. MAE and PSNR values show high consistency among the different test images. 

\begin{figure}[tp!]
\includegraphics[width=\textwidth]{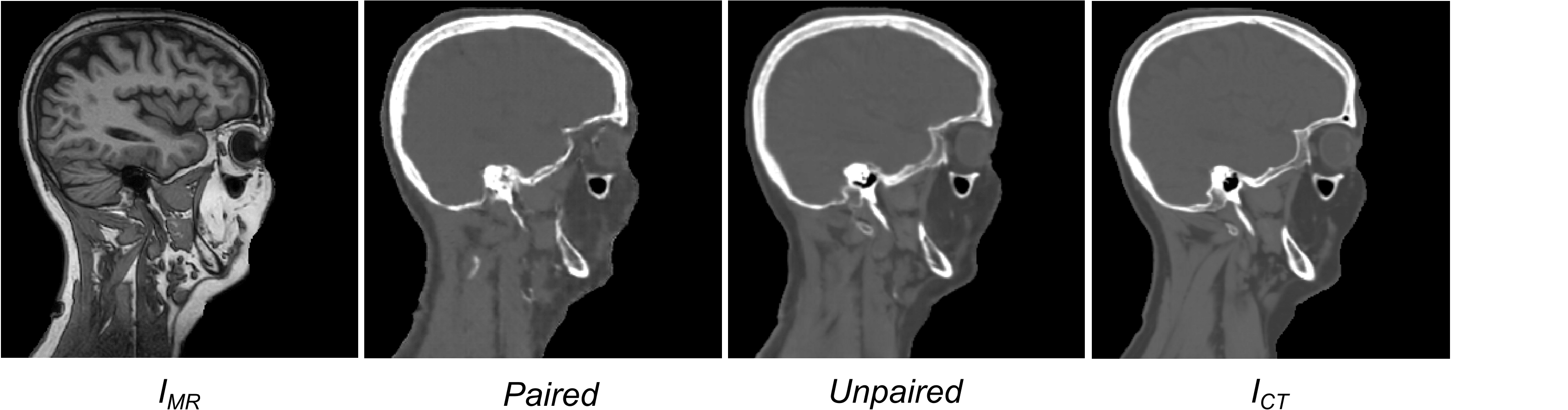}
\caption{\textit{From left to right} Input MR image, synthesized CT image with paired training, synthesized CT image with unpaired training, reference real CT image.}
\label{fig:pairedunpairedexp}
\end{figure}

To compare unpaired training with conventional paired training, an additional synthesis CNN with the same architecture as $Syn_{CT}$ was trained using paired MR and CT image slices. For this, we used the implementation of \cite{Isol16} which, like \cite{Nie16b}, combines voxel-wise loss with adversarial feedback from a discriminator network. This discriminator network had the same architecture as $Dis_{CT}$. A paired t-test on the results in Table \ref{tab:maevalues} showed that agreement with the reference CT images was significantly lower ($p<0.05$) for images obtained using this model than for images obtained using the unpaired model. Fig. \ref{fig:pairedunpairedexp} shows a visual comparison of results obtained with unpaired and paired training data. The image obtained with paired training data is more blurry and contains a high-intensity artifact in the neck.

\begin{table}[t]
\centering
\caption{Mean absolute error (MAE) values in HU and peak-signal-to-noise ratio (PSNR) between synthesized and real CT images when training with paired or unpaired data.}
\label{tab:maevalues}
\begin{tabular}{l|ll|ll}
          & \multicolumn{2}{l|}{MAE} & \multicolumn{2}{l}{PSNR} \\
          & Unpaired     & Paired    & Unpaired     & Paired    \\ \hline
Patient 1 & 70.3             & 86.2          & 31.1             & 29.3          \\
Patient 2 & 76.2             & 98.8          & 32.1             & 30.1          \\
Patient 3 & 75.5             & 96.9          & 32.9             & 30.1          \\
Patient 4 & 75.2             & 86.0          & 32.9             & 31.7          \\
Patient 5 & 72.0             & 81.7          & 32.3             & 31.2          \\
Patient 6 & 73.0             & 87.0          & 32.5             & 30.9          \\
Average $\pm$ SD     & 73.7 $\pm$ 2.3 &   89.4 $\pm$ 6.8  & 32.3 $\pm$ 0.7             & 30.6 $\pm$ 0.9
\end{tabular}
\end{table}

\begin{figure}[tp!]
\includegraphics[width=\textwidth]{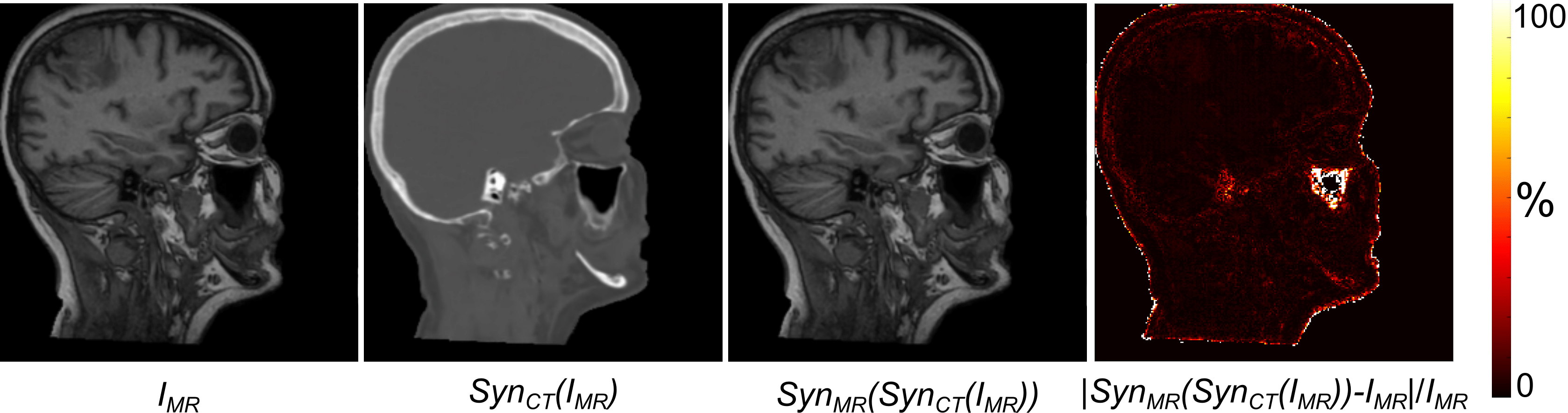}
\caption{\textit{From left to right} Input MR image, synthesized CT image, reconstructed MR image, and relative error between the input and reconstructed MR image.} 
\label{fig:cycleresult}
\end{figure}

During training, cycle consistency is explicitly imposed in both directions. Hence, an MR image that is translated to the CT domain should be successfully translated back to the MR domain. Fig. \ref{fig:cycleresult} shows an MR image, a synthesized CT image and the reconstructed MR image. The difference map shows that although there are errors with respect to the original image, these are very small and homogeneously distributed. Relative differences are largest at the contour of the head and in air, where intensity values are low. The reconstructed MR image is remarkably similar to the original MR image. 

\section{Discussion and Conclusion}
We have shown that a CNN can be trained to synthesize a CT image from an MR image using unpaired and unaligned MR and CT training images. In contrast to previous work, the model learns to synthesize realistically-looking images guided only by the performance of an adversarial discriminator network and  the similarity of back-transformed output images to the original input image.

Quantitative evaluation using an independent test set of six images showed that the average correspondence between synthetic CT and reference CT images was 73.7 $\pm$ 2.3 HU (MAE) and 32.3 $\pm$ 0.7 (PSNR).
In comparison, Nie et al. reported an MAE of 92.5 $\pm$ 13.9 HU and a PSNR of 27.6 $\pm$ 1.3 \cite{Nie16b}, and  Han et al. reported an MAE of 84.8 $\pm$ 17.3 HU \cite{Han17}. However, these studies used different data sets with different anatomical coverage, making a direct comparison infeasible. Furthermore, slight misalignments between reference MR and CT images, and thus between synthesized CT and reference CT, may have a large effect on quantitative evaluation. In future work, we will evaluate the accuracy of synthesized CT images in radiotherapy treatment dose planning. 

Yi et al. showed that a model using cycle consistency for unpaired data can in some cases outperform a GAN-model on paired data \cite{Yi17}. Similarly, we found that in our test data sets, the model trained using unpaired data outperformed the model trained using paired data. Qualitative analysis showed that CT images obtained by the model trained with unpaired data looked more realistic, contained less artifacts and contained less blurring than those obtained by the model trained with paired data. This was reflected in the quantitative analysis. This could be due to misalignment between MR and CT images (Fig. \ref{fig:misalignment}), which is ignored when training with unpaired data.

The results indicate that image synthesis CNNs can be trained using unaligned data. This could have implications for MR-only radiotherapy treatment planning, but also for clinical applications where patients typically receive only one scan of a single anatomical region. In such scenarios, paired data is scarce, but there are many single acquisitions of different modalities. Possible applications are synthesis between MR images acquired at different field strengths \cite{Bahr16}, or between CT images acquired at different dose levels \cite{Wolt17}.

Although the CycleGAN implementation used in the current study was developed for natural images, synthesis was successfully performed in 2D medical images. In future work, we will investigate whether 3D information as present in MR and CT images can further improve performance. Nonetheless, the current results already showed that the synthesis network was able to efficiently translate structures with complex 3D appearance, such as vertebrae and bones. 

The results in this study were obtained using a model that was trained with MR and CT images of the same patients. These images were were rigidly registered to allow a voxel-wise comparison between synthesized CT and reference CT images. We do not expect this registration step to influence training, as training images were provided in a randomized unpaired way, making it unlikely that both an MR image and its registered corresponding CT image were simultaneously shown to the GAN. In addition, images were randomly cropped, which partially cancels the effects of rigid registration. Nevertheless, using images of the same patients in the MR set and the CT set may affect training. The synthesis networks could receive stronger feedback from the discriminator, which would occasionally see the corresponding reference image. In future work, we will extend the training set to investigate if we can similarly train the model with MR and CT images of \textit{disjoint} patient sets. 

\bibliographystyle{splncs03}
\bibliography{wolterink}

\end{document}